\documentclass{article}

\usepackage{arxiv}

\usepackage[utf8]{inputenc} 
\usepackage[T1]{fontenc}    
\usepackage{hyperref}       
\usepackage{url}            
\usepackage{booktabs}       
\usepackage{amsfonts}       
\usepackage{nicefrac}       
\usepackage{microtype}      
\usepackage{lipsum}

\usepackage{times}
\usepackage{epsfig}
\usepackage{graphicx}
\usepackage{amsmath}
\usepackage{amssymb}
  \usepackage{graphicx}
  \graphicspath {{figure/}}
  \usepackage{epstopdf}
  \usepackage{subcaption}

\title{Are spoofs from latent fingerprints a real threat for the best state-of-art liveness detectors?}

\author{Roberto Casula$^1$, Giulia Orrù$^1$, Daniele Angioni$^1$, Xiaoyi Feng$^2$, Gian Luca Marcialis$^1$ and Fabio Roli$^1$
\\$^1$University of Cagliari, Italy\\
$^2$Northwestern Polytechnical University, Xi'an, China\\
{\tt\small \{roberto.casula, giulia.orru, marcialis, roli\}@unica.it, fengxiao@nwpu.edu.cn}
}
\usepackage{tikz}

\newcommand\copyrighttext{%
  \footnotesize \textcopyright 2020 IEEE. Personal use of this material is permitted.
  Permission from IEEE must be obtained for all other uses, in any current or future 
  media, including reprinting/republishing this material for advertising or promotional 
  purposes, creating new collective works, for resale or redistribution to servers or 
  lists, or reuse of any copyrighted component of this work in other works. \\
This paper was accepted for publication in the 25th International Conference on Pattern Recognition (ICPR2020). 
 }
\newcommand\copyrightnotice{%
\begin{tikzpicture}[remember picture,overlay]
\node[anchor=south,yshift=10pt] at (current page.south) {\fbox{\parbox{\dimexpr\textwidth-\fboxsep-\fboxrule\relax}{\copyrighttext}}};
\end{tikzpicture}%
}
\begin{document}
\maketitle
\copyrightnotice
\begin{abstract}
We investigated the threat level of realistic attacks using latent fingerprints against sensors equipped with state-of-art liveness detectors and fingerprint verification systems which integrate such liveness algorithms. To the best of our knowledge, only a previous investigation was done with spoofs from latent prints. In this paper, we focus on using snapshot pictures of latent fingerprints. These pictures provide molds, that allows, after some digital processing, to fabricate high-quality spoofs. Taking a snapshot picture is much simpler  than developing fingerprints left on a surface by magnetic powders and lifting the trace by a tape. What we are interested here is to evaluate preliminary at which extent attacks of the kind can be considered a real threat for state-of-art fingerprint liveness detectors and verification systems. To this aim, we collected a novel data set of live and spoof images fabricated with snapshot pictures of latent fingerprints. This data set provide a set of attacks at the most favourable conditions. We refer to this method and the related data set as ``ScreenSpoof''. Then, we tested with it the performances of the best liveness detection algorithms, namely, the three winners of the LivDet competition. Reported results point out that the ScreenSpoof method is a threat of the same level, in terms of detection and verification errors, than that of attacks using spoofs fabricated with the full consensus of the victim. We think that this is a notable result, never reported in previous work. 

\end{abstract}

\section{Introduction}
A well-known principle in computer security, named “conservative design”, states that if designers of security systems fail to anticipate the capabilities of an adversary major security compromises can occur \cite{joseph_nelson_rubinstein_tygar_2019}. 
Instead, if we are able to anticipate the new capability of adversaries, we can understand the worst-case threat posed by an adversary, and users are less likely to be surprised by an attack by some unanticipated adversary.

Fingerprint spoofing is a well-known presentation attack for biometric recognition systems and previous work proposed different software-based liveness detection algorithms as a defense countermeasure. The most recent algorithms are based on deep learning paradigm that allowed a notable increase of detection accuracy \cite{chugh}.

As to prevent attackers' intention is important in this kind of arms-race problem, we organized the international fingerprint liveness detection competition (LivDet) as the premier forum that shapes the state of the art on spoofs' fabrication techniques and anti-spoofing technology \cite{Orr2019LivDetIA}. Performances of liveness detection algorithms submitted to all the LivDet editions have been mainly assessed with spoof images fabricated by the so called “consensual” method. Spoofs are created by following a rigorous \textit{in vitro} protocol, which includes the careful and controlled pressure of the volunteer’s finger on a plasticine-like material. Once the material is solidified, you have the mold, namely, the 3D fingerprint ''negative'', where locations of ridges and valleys are inverted. Then, the cast material, in a liquid form, is dripped over the mold. Finally, the cast is removed from the mold after the time necessary to its appropriate solidification. The material should have properties similar to those of the human skin: thus, it should be flexible, not too dry, not too moist, and, possibly, wearable or ``stealthy''.

This method is considered a ``worst-case'', namely, the optimal case for the attacker, because it can lead to a very good mold \cite{Marcel}. By the LivDet organization, we experienced that very good molds allow to fabricate spoofs that are likely to deceive the sensor and the matching algorithm used for identity recognition, even if the most recent approaches based on deep learning are adopted \cite{Orr2019LivDetIA}. On the other hand, this kind of attack has a low probability to be executed since it requires the cooperation of the victim; therefore, it can be considered a high-risk attack with a very low probability of execution.
Recently, several videos posted on social networks have shown that very good spoofs can be obtained from latent fingerprints, and such spoofs allowed to crack smartphones of famous brands \cite{forbes,cc,8585605}.
Worth noting, the spoofs lifted from latent prints were also investigated in the 2013 edition of LivDet \cite{livdet2013}. Replicas were obtained by developing the latent traces on a piece of paper with magnetic powders. The LivDet 2013 spoofs have been analyzed in \cite{8608133}, and their effectiveness was quite limited. The development step is a destructive process (it can be done one time only per trace); thus, the quality of the mold entirely depends on the ability of the attacker to perform this process without degrading the quality of the latent mark. Moreover, the scenario where the attacker is able to develop and lift the latent fingerprint of the victim with forensic techniques is not very realistic, as the attacker should have a lot of time and good technical skills.

\begin{figure}[ht]
\centering
\includegraphics[width=.3\textwidth]{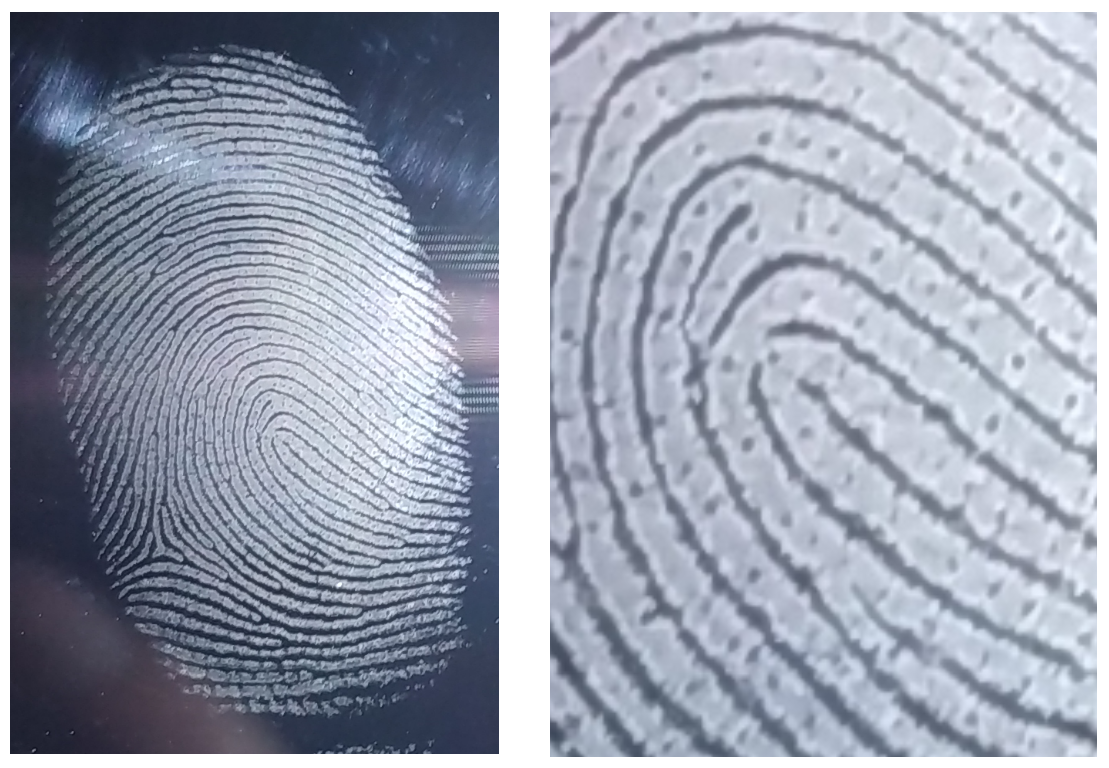}
\caption{Picture of a latent fingerprint on a smartphone screen (left) and details of the pores (right).}
\label{fig:show}
\end{figure}

Recent efforts towards realistic presentation attacks \cite{8585605} led to the observation that latent traces of fingerprints can be captured by a snapshot picture taken with a smartphone; see for example Fig. \ref{fig:show}. This paves the way to unconsensual methods for the fabrication of spoofs where molds are fabricated with the total unawareness of the victim. Taking a snapshot picture requires much less time, efforts and technical skills than the standard forensic procedure that develops the latent trace and puts it on a lifting tape. Finally, according to \cite{8585605}, spoofing by snapshots allows to fabricate fake fingers that pass the identity check of several modern smartphones equipped with fingerprint sensors.
It is easy to see that if spoofs fabricated with snapshot pictures would be able to evade state-of-art fingerprint liveness detectors, we should reconsider current methods to design and test them. The low probability of the ``worst-case'' attack means that this does not have a real impact when assessing the ROC curve of the fingerprint verification system. On the contrary, the probability of the spoofing attack by ``snapshots'' can be expected to be much higher, as this attack is much easier to be executed in the reality. Therefore, such new attack should be taken into serious account when designing anti-spoofing systems and assessing their ROC curves \cite{Chingovska2019}.

Investigating the level of threat of spoofs fabricated with snapshot pictures is the goal of this paper. We use a technique for fingerprint spoofing by snapshot pictures that is much simpler than the one proposed in \cite{8585605}. We called this technique ``ScreenSpoof''. To obtain statistically significant insights, we collected a novel data set of live and spoof fingerprint images fabricated by snapshot pictures. To assess the extent at which such spoofs are a real threat for sensors equipped with the best fingerprint liveness detectors, we assumed that the attacker is able to develop the ``optimal'' latent mark from the smartphone's screen. The ScreenSpoof data set is of the same size than the ones of the leading LivDet competitions. We tested this data set with the three winning algorithms of the last edition of LivDet \cite{Orr2019LivDetIA}. This means that our spoofing attacks were carried out against sensors equipped with the best liveness detection technology currently available.
Our results show that we are at a turning point of the research on fingerprint liveness detection: the detection error and the matching accuracy achieved with our ScreenSpoof technique are comparable with those obtained in the ``worst-case'' scenario where spoofs are fabricated with the full consensus of the victim.
The paper is organized as follows. Section 2 describes the main approaches to spoof fabrication and previous attempts to evaluate the performance of state-of-art fingerprint liveness detectors against replicas from latent traces. Section 3 evaluates the threat level of the ScreenSpoof method presenting the collected data set and reporting the performance of state-of-art fingerprint liveness detectors and matchers taking part at the LivDet 2019 competition. Section 4 closes the paper.
\section{Making fingerprint spoofs: state of the art}
\subsection{Consensual method}
The consensual method for fabrication of fingerprint “spoofs” is considered the ``worst-case'' for anti-spoofing systems as it allows to obtain a perfect replica, or artefact, of a live fingerprint and high-quality spoofs. With the term ``consensual'' method we refer to the basic \textit{molding and casting} method  \cite{matsu}, described in this section, but other more sophisticated 2D and 3D printing techniques can be used \cite{Cao2016HackingMP}. 
The consensual method for \textit{molding and casting} consists of three steps. In the first step, the volunteer pushes the finger into a silicone material to leave the negative impression of her/his biometric trait on a mold. The mold is then filled with a cast material, such as latex, liquid ecoflex or glue. The solidified material is detached from the mold and represents an accurate copy of the real fingerprint and can be used to execute a presentation attack against a fingerprint recognition system (also equipped with a liveness detector).

Although this method may achieve the goal of evading a fingerprint verification system, it is not realistic as it requires the consensus of the victim. It can be successful only in the hypothetical case that the target user is a partner in crime of the attacker, or the attacker is so skillful that the victim does not realize that a copy of her/his fingerprint has been stolen, for example, by an accidental pressure on a wax or plasticine surface. Even in this favourable case, the mold could not be useful to provide a good replica of the victim's fingerprint. Therefore, if this method is optimal \textit{in vitro}, it is not easy to execute successfully in a real context.  

\subsection{Unconsensual method}
The non-consensual method is the most realistic and ``dangerous'' method by which an attacker can fabricate spoofs and evade a fingerprint recognition system. In fact, it does not require the victim's consensus and collaboration. Usually, a latent fingerprint is taken from a smooth or nonporous surface through magnetic powders and digitized through a scanner or a photograph.
The cast material is then applied to a transparent sheet on which the negative latent has been printed. This implies more or less complex digital processing to make useful the fingerprint negative by a printer. Another way to create the spoof is to etch the fingerprint negative onto a printed circuit board and then drip the material.

The fabrication of spoofs from latent prints is usually difficult because it requires great skills, specialized equipment and time enough.
\subsubsection{Previous work on spoofing by latent prints}
A first assessment of the level of threat of attacks with spoofs fabricated by the unconsensual method was described in the LivDet 2013 paper \cite{livdet2013}. Two data sets were created using latent fingerprints collected by dusting a sheet of paper with fingerprint powder. 
The developed fingerprint was then photographed and digitized, and the negative image was printed on a transparency sheet. Using this printed image as mold, spoofs were created with different materials (gelatine, latex, ecoflex, modasil and wood glue).

The LivDet 2013 results showed that this replication technique is only partially effective. In fact, the spoofs were apparently easy to detect due to the low Equal Error Rate (EER) obtained.
In particular, the three winning algorithms reported performances over 95\% with the ``unconsensual'' data sets and much lower performances with the ``consensual'' data sets.

This result is confirmed by \cite{GHIANI2017110} that compared the LivDet 2013 average results achieved with Biometrika and Italdata sensors with those of the LivDet 2011 edition in which these two sensors were used with the consensual method. The authors pointed out that in the 2013 edition, the error rate with these two data sets was 20\% less than the one of the previous edition (from 30\% to around 10\% in terms of average detection rate).
Although algorithms and participants were different, the above figure clearly shows that, in 2011, spoofs fabricated from latent prints were far from being a real threat for fingerprint liveness detection systems.
\subsubsection{Spoofing by snapshot pictures of latent prints}
The reflective surfaces easily get dirty with the human skin residues consisting of secretions of the eccrine, sebaceous and apocrine glands. Recently, it has been shown that latent fingerprints left on smartphones and touchscreen-equipped mobile devices are easy to identify without the use of specialized equipment and can be used to evade a fingerprint authentication system. In 2013, the hackers of the Chaos Computer Club (CCC) \cite{cc} broke the Apple TouchID using a photo-sensitive PCB mold created by taking a picture or scanning a latent fingerprint on a smartphone screen.

The effectiveness of this new spoofing technique was confirmed later by \cite{8585605}, in which the authors simulated an attack against five different smartphones by collectiong the latent fingerprints from the device screens.
In particular, the technique used by \cite{8585605} was based on a scanner app to take the picture with flash of the latent fingerprint. The acquired image was then improved by removing the background and other adjustments.
Finally, the mold was created from the pre-processed image by  PCB techniques and four materials were used to fabricate spoofs (Play-Doh, gelatin, latex with sprayed graphite powder and white glue with sprayed graphite powder).

The target devices were vulnerable at the attack, reporting an average  Impostor Attack Presentation Match Rate (IAPMR) of 9\%.
Three out of four of the materials used to fabricate spoofs were effective against smartphone sensors, although there was a substantial performance difference among the targeted victims.

This previous work did not check if a modern liveness detection system could identify this attack and whether this technique for spoofs fabrication is better or worst than the techniques used at state-of-art.
In the following sections, we investigate the level of threat of an attack using spoofs fabricated from snapshot pictures of latent prints.

\section{Making spoofs by snapshot: a real threat?}
\subsection{The ScreenSpoof data set}

\begin{figure}[ht]
\centering
\includegraphics[width=.35\textwidth]{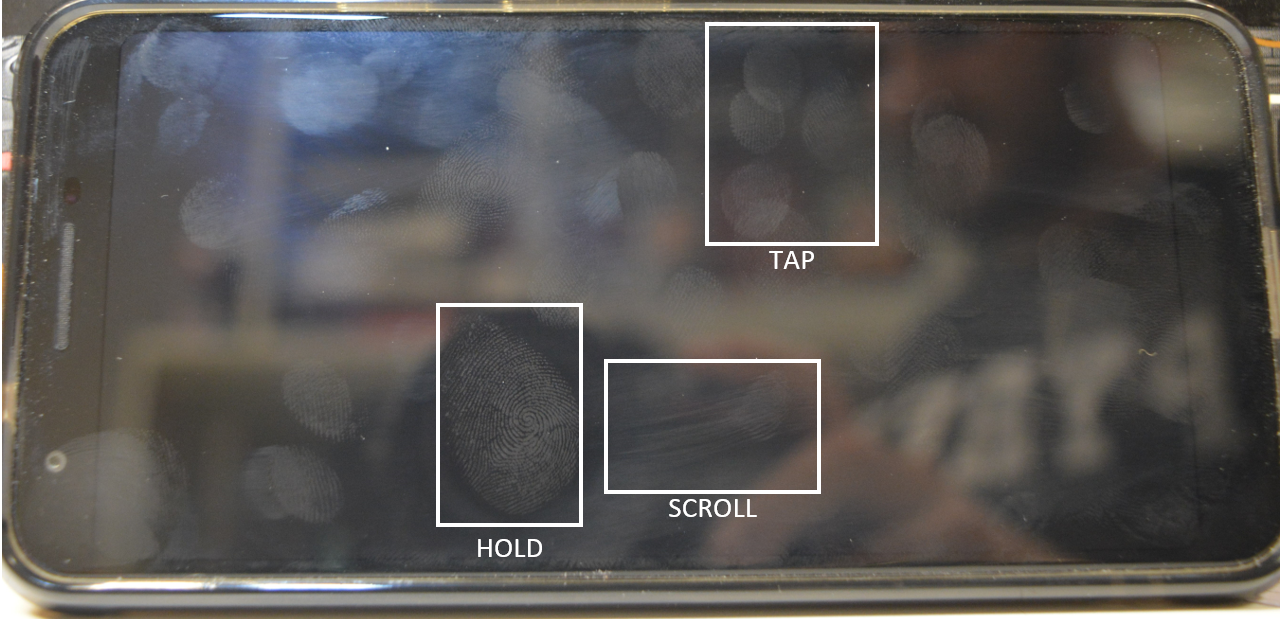}
\caption{Typology of impressions on screens based on usage.}
\label{fig:wildscreen}
\end{figure}

\begin{figure}[ht]
\centering
\begin{subfigure}[b]{0.49\linewidth}
\centering
\includegraphics[width=1\textwidth]{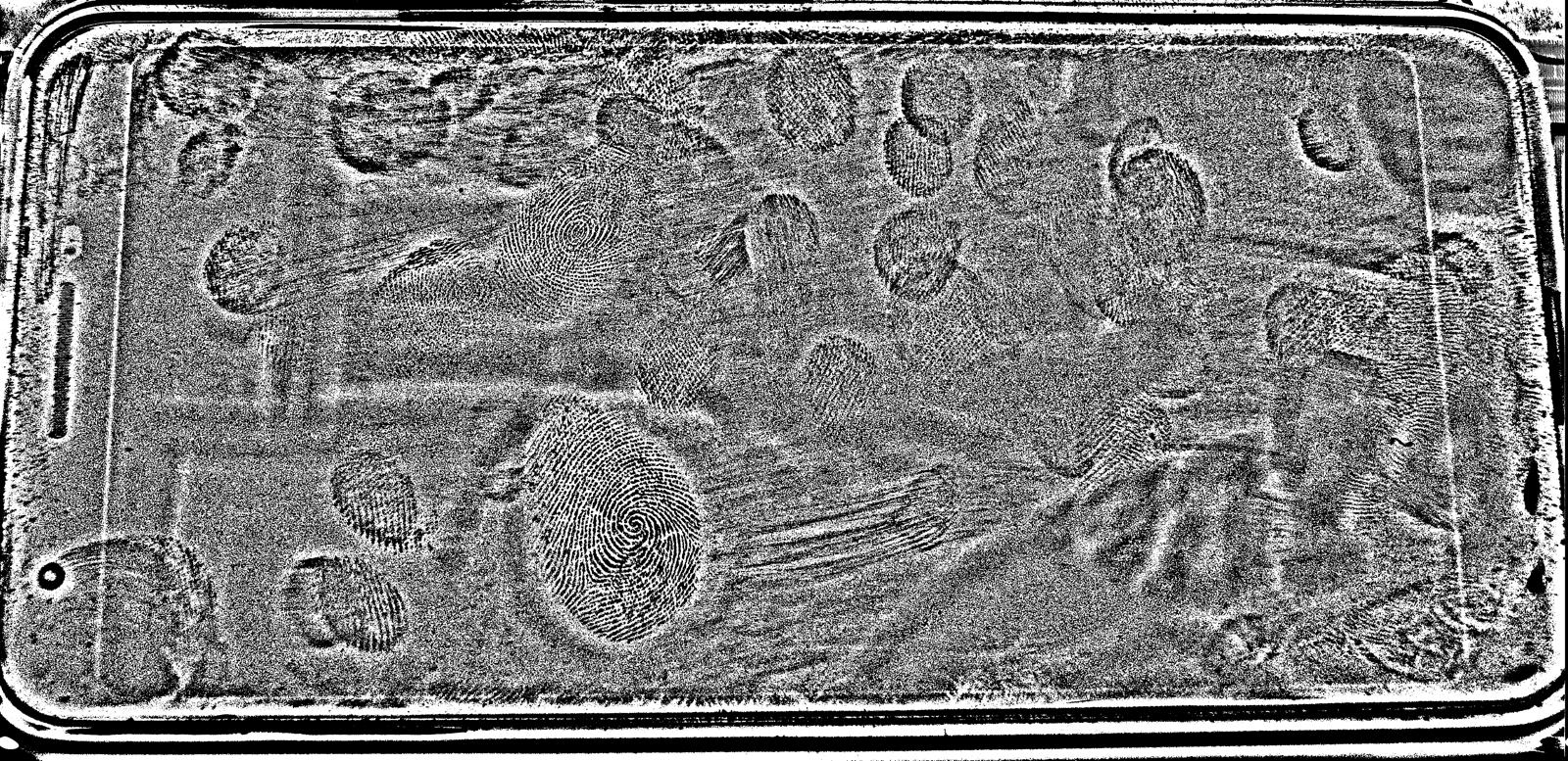}
\end{subfigure}
\begin{subfigure}[b]{0.49\linewidth}
\centering
{\includegraphics[width=0.4\textwidth]{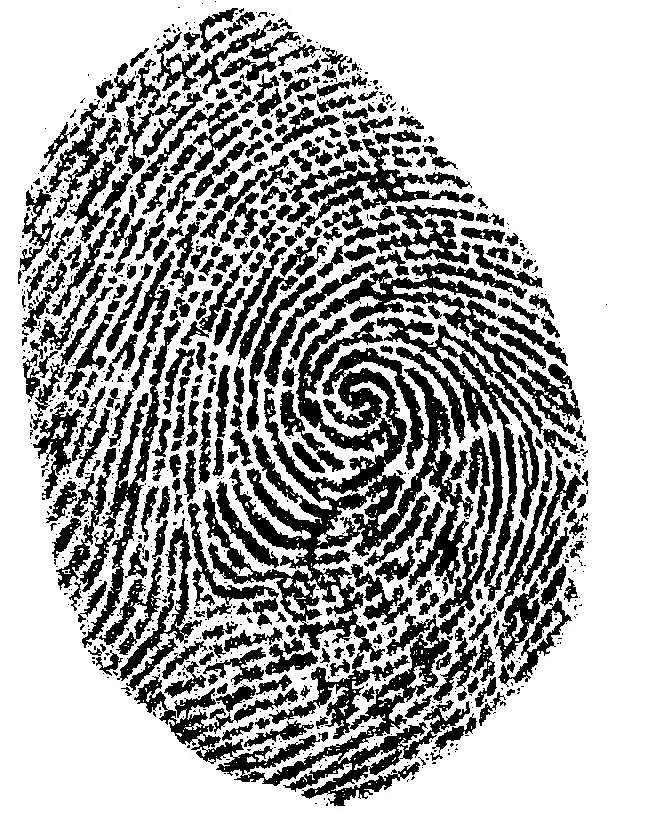}}
\end{subfigure}
\caption{Development by pre-processing of a fingerprint acquired ``in the wild''.}
\label{fig:wildprepr}
\end{figure}

To assess the level of threat of a realistic approach that steals latent fingerprints without the consensus of the victim, we collected a novel data set called ``Stealthy Spoofing from Smartphone'' or ``ScreenSpoof'' data set and made up of live and fake fingerprint images. In the following, we refer to both spoofing methods and data by the term ``ScreenSpoof''.
We were inspired by the technique presented in \cite{8585605} but we further simplified it to achieve an even more realistic attack. This technique is based on taking a snapshot of the smartphone screen, where it is almost sure that a good set of possible latent ``molds'' can be found.
\\
It is worth noting that a latent fingerprint is often characterized by (a) a noisy/complex
background, (b) partial friction ridge information, and (c) poor friction ridge clarity \cite{latentjain}. In particular, the fingerprints left on the screens are due to the actual use of the device, as in the case of tap and scroll, which almost always lead to partial and overlapping fingerprints. However, holding the device in hand causes non-overlapping impressions with complete and clear friction ridge information. Fig. \ref{fig:wildscreen} exemplifies the above cases. In the last one, an appropriate pre-processing of the image allows to obtain a useful mold, as shown in Fig. \ref{fig:wildprepr}. 
It is hence obvious that the attacker has the time to choose the best one among latent prints visually detected, once she/he took the snapshot.  
\\
On the basis of these observations, our ``ScreenSpoof'' data set was acquired \textit{in vitro}, with the collaboration of the users. As shown in Fig. \ref{fig:wildscreen}, the same type of candidate images can be obtained ``in the wild'', but with much more expense of time and efforts. Leaving the problem of this task to a next publication, we dealt with the ``worst case'', where the attacker was able to develop a full and complete latent print left on the screen by the incautious user.

\begin{figure*}[ht]
\centering{\includegraphics[width=.8\textwidth]{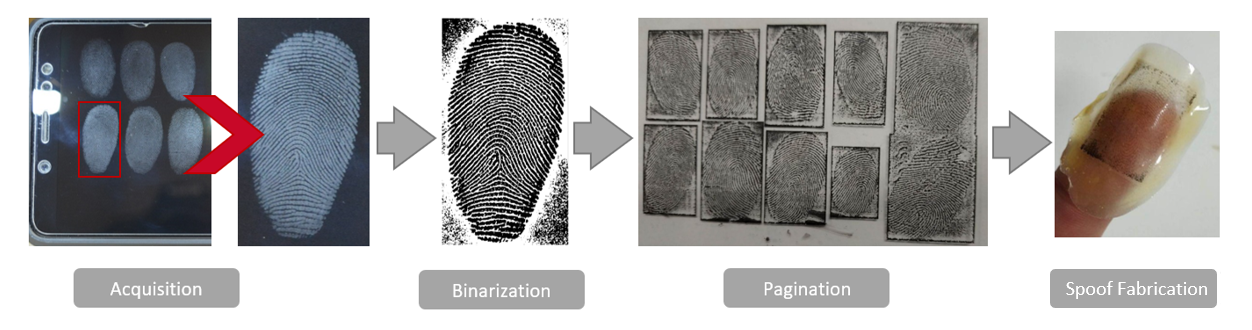}}
\caption{The four collection steps of the ScreenSpoof data set: the latent fingerprint is (1) photographed from the device screen, (2) pre-processed, (3) printed on a transparent sheet and (4) spoof is fabricated by pouring a material over the sheet.}
\label{fig:schemaSLS}
\end{figure*}

The main collection phases of our data set are (Fig. \ref{fig:schemaSLS}):
\begin{enumerate}
    \item Acquisition: once the screen has been carefully cleaned, the user is asked to place index, middle and ring fingers of the right hand in the upper part and the same fingers of the left hand in the lower part; then the smartphone is placed on a vertical support in order to take a high-resolution photo of all six fingerprints. The same procedure is performed for the thumb and little finger of both hands. Users have not been asked to wash their hands, nor to use moisturizers, the quality of the impression depends on the natural dryness and dirtyness of the skin during the acquisition.
    \item Binarization: the RGB image is converted to grayscale, inverted and image contrast between ridges and valleys increased.
    \item Pagination: the photo is cropped and scaled with respect to the size of the real fingerprint, using reference points on the screen. 
    \item Fake fabrication: the resulting ``negative'' image is printed on a transparent sheet. The creation of spoofs, starting from the transparent sheet print, coincides with the traditional non-consensual method described above: the sheet is used as mold.
    Different cast materials are dripped over several prints of the mold. Once dried, spoofs are removed from the sheet and captured with the sensors.
\end{enumerate} 
The ScreenSpoof data set consists of 30 people: five acquisitions were made per finger, reaching a total of 1500 images per sensor. Spoofs were fabricated with three different materials (Mix 1, Mix 2 and body double) and acquired twice for a total of 1800 images (600 per material) per sensor. \\
All the acquisitions were carried out with two sensors, the Green Bit and Digital Persona sensors (Table \ref{tab:sensors}), already used in the LivDet 2019 \cite{Orr2019LivDetIA} so that we are able to compare the state-of-art consensual method with our ScreenSpoof method. 

Fingerprints quality varies from user to user based on the dryness of the skin and the cosmetics used (e.g. individuals who use foundation and often touch their facial skin leave clear impressions on the screen).
\begin{table}[ht]
\centering
\resizebox{0.6\textwidth}{!}{%
\begin{tabular}[ht]{ | l | c | c | c | c | c |}
\hline
\textbf{Scanner} & \textbf{Model} & \textbf{DPI} & \textbf{Image Size} &\textbf{Type} \\ \hline
Green Bit & DactyScan84C & 500 & 500x500 & Optical \\ \hline
Digital Persona& U.are.U 5160 & 500  &252x324 & Optical \\ \hline
\end{tabular}}
\caption{Green Bit and Digital Persona device features.}
\label{tab:sensors}
\end{table}
\begin{table*}[t]
\centering
\resizebox{0.9\textwidth}{!}{%
\begin{tabular}{|c||c|c|c|c||c|c|c|c||}
\hline
                 & \multicolumn{4}{c||}{\textbf{Test ScreenSpoof}}      & \multicolumn{4}{c||}{\textbf{Test LivDet 2019}}\\ \hline
\textbf{Dataset} & Live & Mix 1 & Mix 2 & Body Double & Live & Mix 1 & Mix 2 & Liquid Ecoflex\\ \hline
Green Bit        &1500&600&600&600&1020&408&408&408\\ \hline
Digital Persona  &1500&600&600&600&1019&408&408&408\\ \hline
\end{tabular}%
}
\caption{Details of the two test sets used to compare the consensual method with our ScreenSpoof. The algorithms had previously been trained with the LivDet 2019 train set (using the consensual method).}
\label{tab:datasetdComposition}
\end{table*}
\subsection{Threat evaluation}
As we explained before, the ScreenSpoof fabrication technique is realistic and easy to use. Therefore, it is important to evaluate how much it is also effective and it could be a real threat against state-of-art liveness detectors.

\begin{figure*}[ht]
\begin{subfigure}[b]{1\linewidth}
\centering{\includegraphics[width=0.8\textwidth]{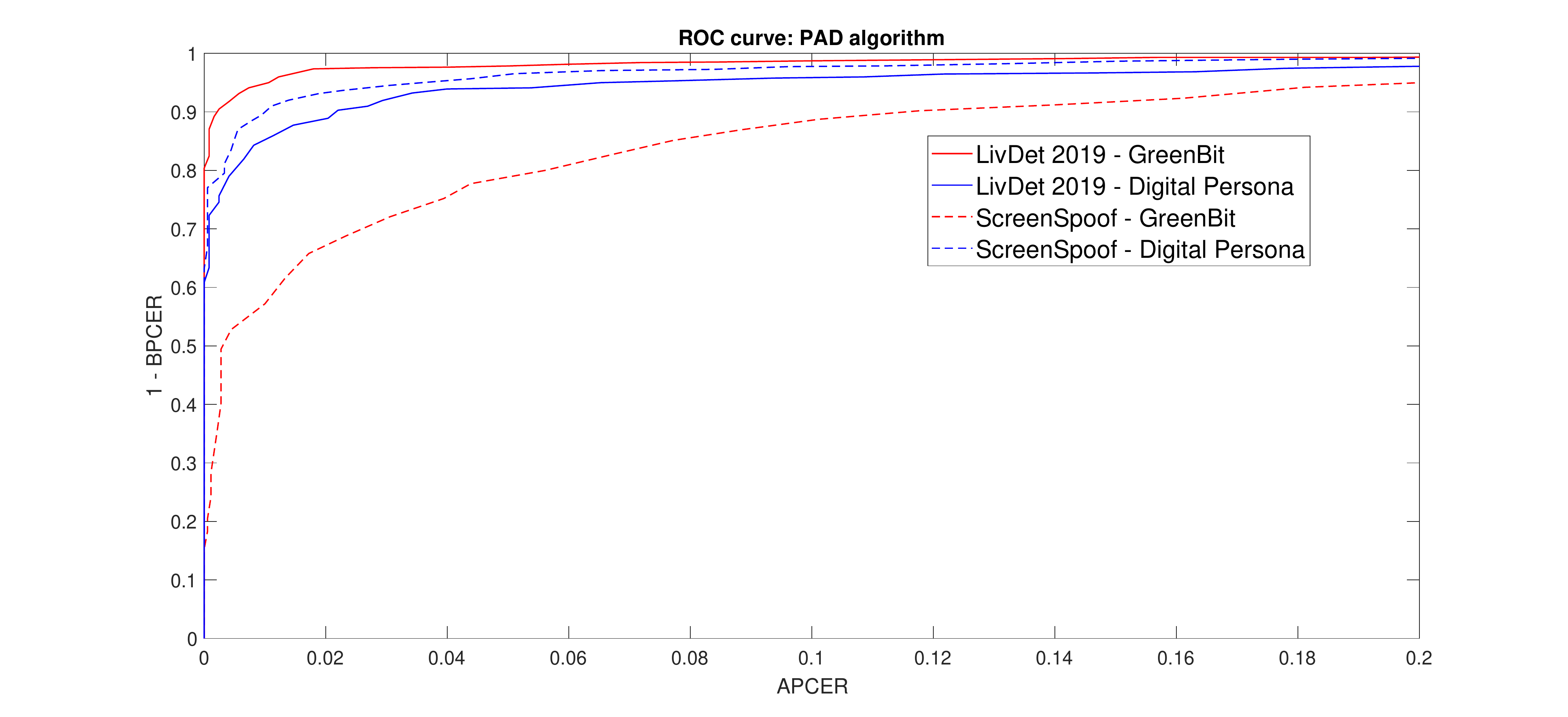}}
\subcaption{PAD ROC}
\end{subfigure}\\
\begin{subfigure}[b]{1\linewidth}
\centering\includegraphics[width=0.8\textwidth]{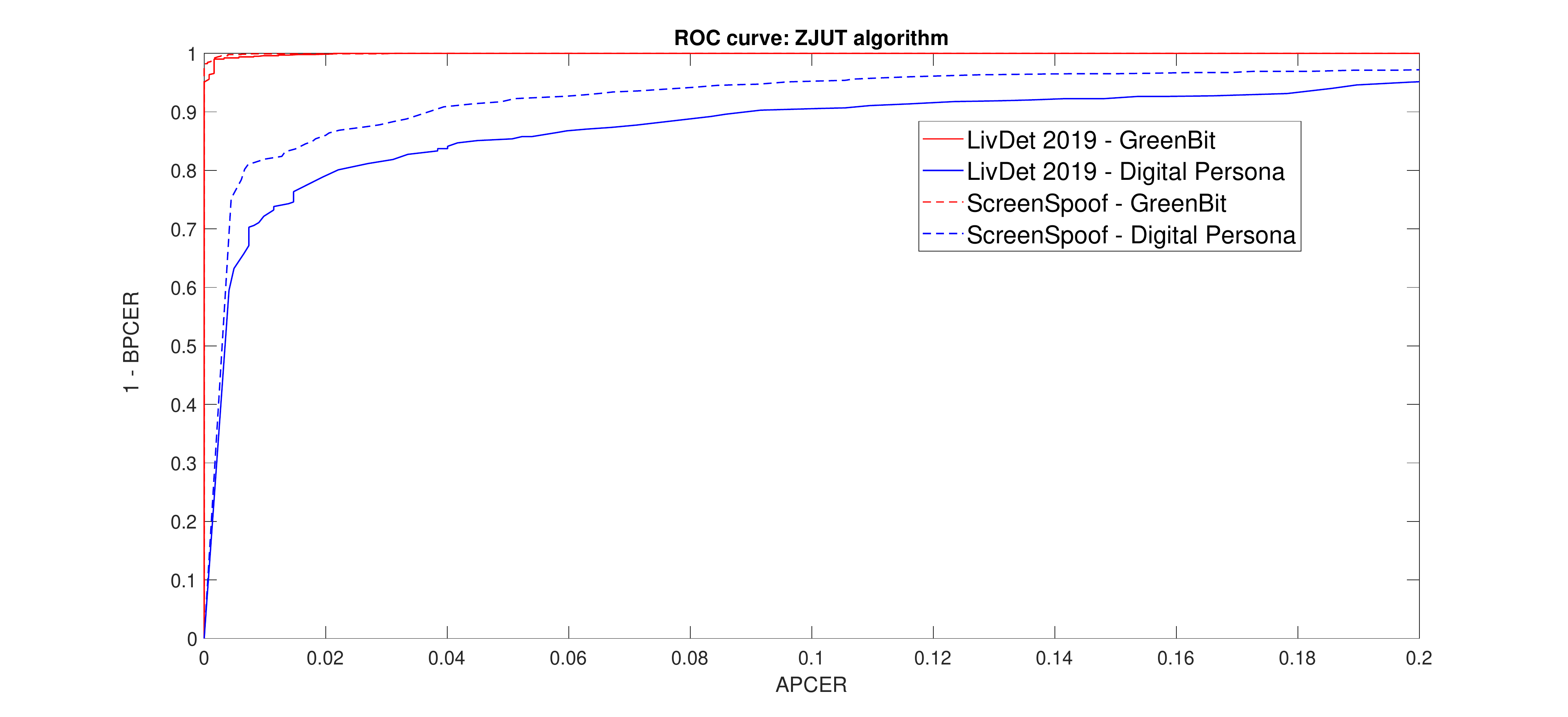}
\subcaption{ZJUT ROC}
\end{subfigure}\\
\begin{subfigure}[b]{1\linewidth}
\centering\includegraphics[width=0.8\textwidth]{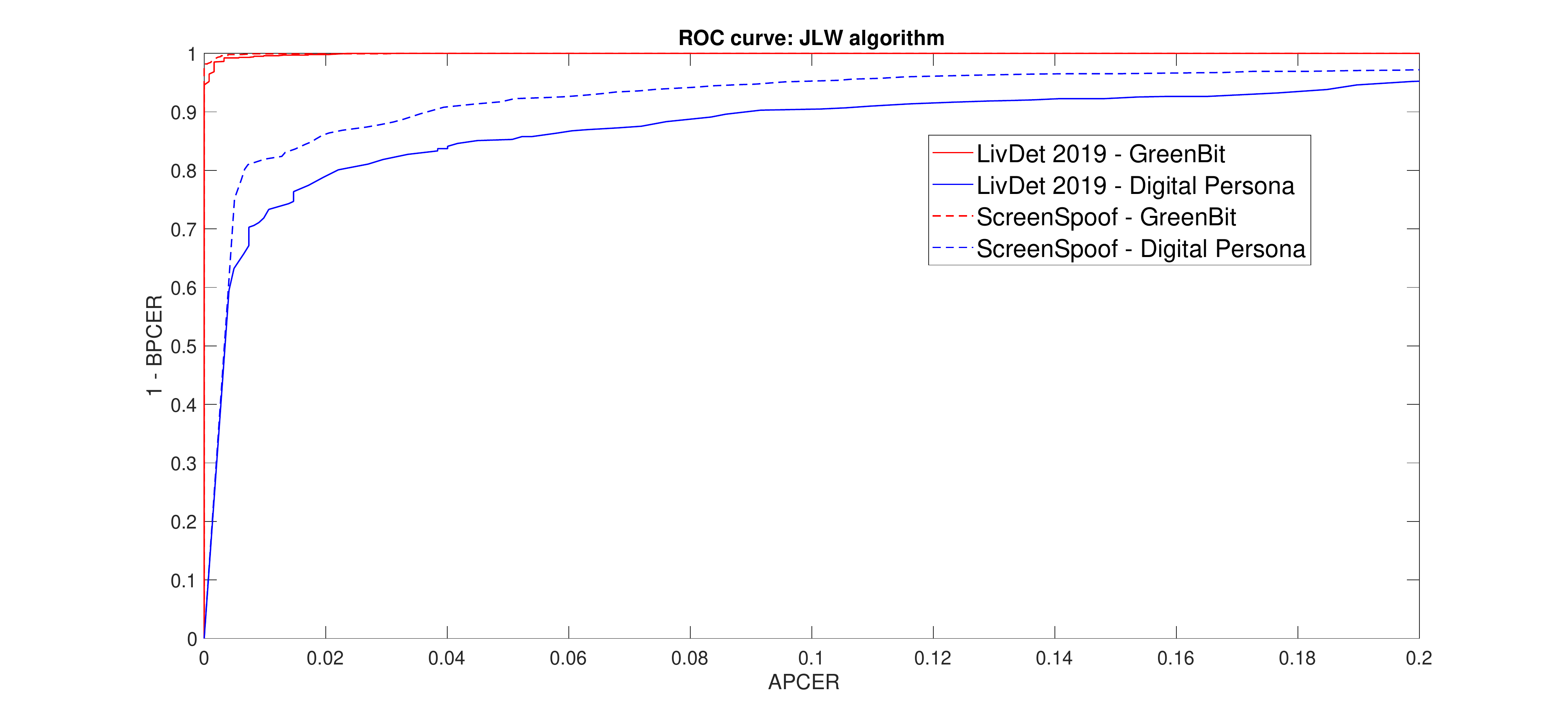}
\subcaption{JLW ROC}
\end{subfigure}
\caption{ROC curves of the three winning algorithms of LivDet 2019 for the two sensors: the blue lines indicate the Green Bit sensor and the red lines indicate the  Digital Persona sensor. The continuous lines indicate the ROCs calculated using the LivDet 2019 test set, the dotted ones indicate the ScreenSpoof test set.}
\label{fig:roc}
\end{figure*}

To this aim, we tested the three winning algorithms submitted to the last edition of the LivDet competition \cite{Orr2019LivDetIA}, namely, the algorithms named PAD, ZJUT, and JLW, with our ScreenSpoof data set.
These three algorithms have different features:
PAD is a presentation attack detection system based on handcrafted features, ZJUT and JLW are fingerprint verification systems also equipped with a presentation attack detection module based on deep learning techniques.
In particular, the PAD method uses a combination of local and global feature to characterize fingerprints \cite{gonzlezsoler2019fingerprint}. The ZJUT algorithm uses a residual convolutional neural network, called Slim-ResCNN \cite{8756285}. We have no further information on the JLW method. In particular, we used the ZJUT\_Det\_A and JLWs versions described in \cite{Orr2019LivDetIA}.
All algorithms have been pre-trained with the LivDet 2019 training set \cite{Orr2019LivDetIA}.
We analyzed liveness detection performances for all three algorithms, and recognition performances for ZJUT and JLW, using the experimental protocol of the LivDet 2019 competition on LivDet 2019 testing set and on the ScreenSpoof data set. Therefore, the experiments are cross-material and cross-database, in line with the typical fingerprint presentation attack detection evaluation protocols, for which the absence of knowledge of the type of spoofing attack is a fundamental point.
Figure \ref{fig:roc} shows the ROC curves of the three algorithms in terms of liveness detection performance, the Bona fide Presentation Classification Error Rate (BPCER) and the Attack Presentation Classification Error Rate (APCER). For each plot, we compare performances of spoofs fabricated with the consensual method (``worst-case'' scenario) with the ones of spoofs fabricated with our technique using snapshot pictures, for Green Bit sensor (in blue) and Digital Persona sensor (in red).
The continuous lines indicate the ROCs calculated with the LivDet 2019 test set, the dotted ones refer to the ScreenSpoof test set.
It is worth noting that our spoofs by snapshot pictures provide performances very close to the ones of the consensual method.
Our spoofs are sometimes even more difficult to detect than the ``consensual'' spoofs. This is a notable result never reported before.

BPCER@1\%APCER and APCER@1\%BPCER values reported in Table \ref{table:BPCERAPCER} confirms the effectiveness of our spoofs by snapshot pictures and, therefore, the high level of threat of this kind of spoofing attack.
In particular, the liveness detection algorithm tested with images acquired with the Green Bit sensor shows a very high drop of the performance.

\begin{figure}[ht]
\centering
\begin{subfigure}[b]{0.33\linewidth}
\centering
\includegraphics[width=0.3\textwidth]{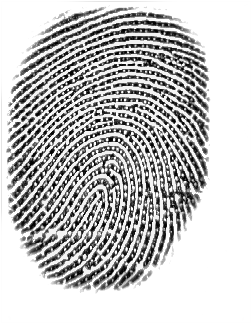}
\subcaption{Live}
\end{subfigure}
\begin{subfigure}[b]{0.33\linewidth}
\centering
\includegraphics[width=0.3\textwidth]{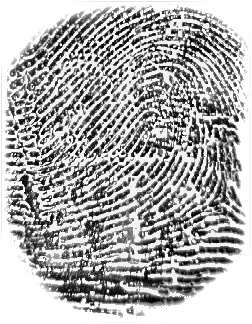}
\subcaption{Consensual}
\end{subfigure}
\begin{subfigure}[b]{0.33\linewidth}
\centering
{\includegraphics[width=0.3\textwidth]{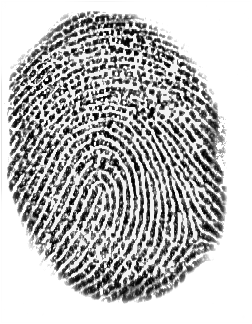}}
\subcaption{ScreenSpoof}
\end{subfigure}
\caption{Example of live and spoof images adopted in the test stage of the LivDet 2019 algorithms.}
\label{fig:finger}
\end{figure}

Some hypotheses can be done to explain the reasons of the noticeable performance of the spoofs obtained by our method. Just to give a first motivation, we show in Fig. \ref{fig:finger} some example images of the spoofs obtained by our ScreenSpoof and consensual methods and the related live images. Images look  similar by visual and optical microscope inspections: ridge and valleys are clearly replicated, and pores are visible even in the ScreenSpoof-based image. This qualitative evaluation may lead to the conclusion that both methods are able to effectively replicate the main features of the ``stolen'' fingerprint. 
On the other hand, further investigations are necessary to support better and clarify the cases where our ScreenSpoof method clearly outperforms the consensual one.


Tables \ref{table:integGB} and \ref{table:integDP} show the performance of the two integrated systems, ZJUT and JLW, respectively for GreenBit and Digital Persona sensors. The values of False Match Rate (FMR), False Non-Match Rate (FNMR) and Impostor Attack Presentation Match Rate (IAPMR) \cite{Chingovska2019} are computed as follows. If the integrated match score between the submitted image and the template is more than 0.5, according to the LivDet 2019 competition rules, the image is verified as belonging to the claimed identity. Otherwise it is rejected as zero-effort (impostor) or presentation attack.

Between the ScreenSpoof and LivDet results, there are some performance fluctuations. Genuine users of the ScreenSpoof data set are better recognized (FMR) while it is easier to correctly classify the zero-effort impostors of the LivDet data set (FNMR). Since in both cases, these images have been produced from live fingerprints, these fluctuations cannot be due to the spoof fabrication technique and derive from the response of the presentation attack detector integrated on the matcher. It should be noted that we do not know the matchers' details and the rule of integration with the presentation attack detector.

The IAPRM value is a measurement of how ``dangerous'' these spoofs can be for an integrated system (i.e., a fingerprint verification system integrated with a liveness detector). It is worth noting that the variations of IAPMR values differ substantially on the basis of the sensor used. For the Green Bit sensor, ScreenSpoof attacks are less effective than the consensual ones. The PAs created with the ScreenSpoof technique and acquired with the Digital Persona sensor are instead very difficult to be classified correctly.
Since the Digital Persona data set was the most ``difficult'' one in LivDet 2019, the evidence of an attack even harder to be detected points out the need for a better understanding of the features replicated by the ScreenSpoof method.    

\begin{table*}[!ht]
\centering
\resizebox{1\textwidth}{!}{%
\begin{tabular}{c|c|c|c|c|c|c|c|c|}
\cline{2-9}
                                    & \multicolumn{4}{c|}{\textbf{GreenBit}}                                              & \multicolumn{4}{c|}{\textbf{Digital Persona}}                                       \\ \cline{2-9} 
                                    & \multicolumn{2}{c|}{\textbf{BPCER@1\%APCER}} & \multicolumn{2}{c|}{\textbf{APCER@1\%BPCER}} & \multicolumn{2}{c|}{\textbf{BPCER@1\%APCER}} & \multicolumn{2}{c|}{\textbf{APCER@1\%BPCER}} \\ \cline{2-9} 
                                    & \textbf{LivDet 2019}    & \textbf{ScreenSpoof}   & \textbf{LivDet 2019}    & \textbf{ScreenSpoof}   & \textbf{LivDet 2019}    & \textbf{ScreenSpoof}   & \textbf{LivDet 2019}    & \textbf{ScreenSpoof}   \\ \hline
\multicolumn{1}{|c|}{\textbf{PAD} \cite{gonzlezsoler2019fingerprint}}  & 5.00\%                  & 40.80\%        & 14.22\%                 & 48.39\%        & 14.03\%                 & 8.93\%         & 40.95\%                 & 20.61\%        \\ \hline
\multicolumn{1}{|c|}{\textbf{ZJUT} \cite{8756285}} & 0.39\%                  & 0.07\%         & 0.16\%                  & 0.17\%         & 26.76\%                 & 18.07\%        & 55.68\%                 & 27.56\%        \\ \hline
\multicolumn{1}{|c|}{\textbf{JLW}}  & 0.39\%                  & 0.07\%         & 0.33\%                  & 0.17\%         & 26.67\%                 & 18.07\%        & 55.60\%                 & 27.67\%        \\ \hline
\end{tabular}}
\caption{Comparison of BPCER@1\%BPCER and APCER@1\%BPCER for the three most accurate liveness detectors of the LivDet 2019 competition using a consensual test set (LivDet 2019 test) and a unconsensual test set (ScreenSpoof data set).}
\label{table:BPCERAPCER}
\end{table*}

\begin{table*}[!ht]
\centering
\resizebox{0.7\textwidth}{!}{%
\begin{tabular}{c|c|c|c|c|c|c|}
\cline{2-7}
                                    & \multicolumn{6}{c|}{\textbf{GreenBit}}                                                                          \\ \cline{2-7} 
                                    & \multicolumn{2}{c|}{\textbf{FMR}}   & \multicolumn{2}{c|}{\textbf{FNMR}}  & \multicolumn{2}{c|}{\textbf{IAPMR}} \\ \cline{2-7} 
                                    & \textbf{LivDet 2019} & \textbf{ScreenSpoof} & \textbf{LivDet 2019} & \textbf{ScreenSpoof} & \textbf{LivDet 2019} & \textbf{ScreenSpoof} \\ \hline
\multicolumn{1}{|c|}{\textbf{ZJUT}} & 0.49\%               & 0.33\%       & 5.48\%               & 7.31\%       & 2.60\%               & 1.33\%       \\ \hline
\multicolumn{1}{|c|}{\textbf{JLW}}  & 0.76\%               & 0.58\%       & 0.03\%               & 0.46\%       & 2.65\%               & 1.44\%       \\ \hline
\end{tabular}}
\caption{Evaluation of integrated algorithms (matching algorithms and liveness detectors) under presentation attacks using a consensual test set (LivDet 2019 test) and a unconsensual test set (ScreenSpoof data set) for the GreenBit sensor.}
\label{table:integGB}
\end{table*}

\begin{table*}[!ht]
\centering
\resizebox{0.7\textwidth}{!}{%
\begin{tabular}{c|c|c|c|c|c|c|}
\cline{2-7}
                                    & \multicolumn{6}{c|}{\textbf{Digital Persona}}                                                                   \\ \cline{2-7} 
                                    & \multicolumn{2}{c|}{\textbf{FMR}}   & \multicolumn{2}{c|}{\textbf{FNMR}}  & \multicolumn{2}{c|}{\textbf{IAPMR}} \\ \cline{2-7} 
                                    & \textbf{LivDet 2019} & \textbf{ScreenSpoof} & \textbf{LivDet 2019} & \textbf{ScreenSpoof} & \textbf{LivDet 2019} & \textbf{ScreenSpoof} \\ \hline
\multicolumn{1}{|c|}{\textbf{ZJUT}} & 6.93\%               & 2.68\%       & 8.79\%               & 12.38\%      & 8.79\%               & 22.87\%      \\ \hline
\multicolumn{1}{|c|}{\textbf{JLW}}  & 11.52\%              & 6.10\%       & 0.19\%               & 0.59\%       & 8.73\%               & 14.20\%      \\ \hline
\end{tabular}}
\caption{Evaluation of integrated algorithms under presentation attacks using a consensual test set (LivDet 2019 test) and a unconsensual test set (ScreenSpoof data set) for the Digital Persona sensor.}
\label{table:integDP}
\end{table*}

\section{Conclusions}

The design and test of fingerprint liveness detection systems was made so far by adopting the consensual technique to fabricate high-quality spoofs. However, the effort to provide very good quality molds and spoofs is not easy to provide \textit{in vitro}, and can be even more difficult in real scenarios, as we experienced during the organization of the editions of LivDet. Therefore, a presentation attack of that kind is unlikely to be executed. 

In this paper, we tried to anticipate attackers of fingerprint recognition systems by proposing a realistic technique that uses snapshot pictures of latent fingerprints to fabricate high quality spoofs, which we called ScreenSpoof. Despite this data set is still made up of spoof images collected by assuming the ability of the attacker in developing the best latent mark, our results point out that this kind of attack is a clear and present threat of the same level, in terms of detection and verification errors, than that of attacks using spoofs fabricated with the full consensus of the victim. Moreover, the apparent effort to fabricate fingerprint artefacts by the ScreenSpoof method is much less than the unconsensual approaches based on forensic techniques, and can be carried out by fooling the victim.
\\
This is a notable result, never reported in previous work: attackers have a realistic chance to evade the best liveness detection technology currently available. This was confirmed by the performance of the three winning algorithms of the LivDet 2019 competition. 

Our future work will be focused on large scale comparisons among unconsensual spoofing attacks. In particular, we will check the extent at which an attacker can obtain “stealthy” molds in-the-wild in terms of the latent's size and quality.

\section*{Acknowledgements}
We would like to thank the three winners of the LivDet 2019 competition, Airel Pérez-Suárez, Yongliang Zhang and Jinglian Wen, for allowing us the use of their PAD algorithms.


\bibliographystyle{IEEEtran}
\bibliography{template}

\begin{thebibliography}{10}
\providecommand{\url}[1]{#1}
\csname url@samestyle\endcsname
\providecommand{\newblock}{\relax}
\providecommand{\bibinfo}[2]{#2}
\providecommand{\BIBentrySTDinterwordspacing}{\spaceskip=0pt\relax}
\providecommand{\BIBentryALTinterwordstretchfactor}{4}
\providecommand{\BIBentryALTinterwordspacing}{\spaceskip=\fontdimen2\font plus
\BIBentryALTinterwordstretchfactor\fontdimen3\font minus
  \fontdimen4\font\relax}
\providecommand{\BIBforeignlanguage}[2]{{%
\expandafter\ifx\csname l@#1\endcsname\relax
\typeout{** WARNING: IEEEtran.bst: No hyphenation pattern has been}%
\typeout{** loaded for the language `#1'. Using the pattern for}%
\typeout{** the default language instead.}%
\else
\language=\csname l@#1\endcsname
\fi
#2}}
\providecommand{\BIBdecl}{\relax}
\BIBdecl

\bibitem{joseph_nelson_rubinstein_tygar_2019}
A.~D. Joseph, B.~Nelson, B.~I.~P. Rubinstein, and J.~D. Tygar,
  \emph{Adversarial Machine Learning}.\hskip 1em plus 0.5em minus 0.4em\relax
  Cambridge University Press, 2019.

\bibitem{chugh}
T.~{Chugh}, K.~{Cao}, and A.~K. {Jain}, ``Fingerprint spoof buster: Use of
  minutiae-centered patches,'' \emph{IEEE Transactions on Information Forensics
  and Security}, vol.~13, no.~9, pp. 2190--2202, Sep. 2018.

\bibitem{Orr2019LivDetIA}
G.~Orr{\`u}, R.~Casula, P.~Tuveri, C.~Bazzoni, G.~Dessalvi, M.~Micheletto,
  L.~Ghiani, and G.~L. Marcialis, ``Livdet in action - fingerprint liveness
  detection competition 2019,'' \emph{ArXiv}, vol. abs/1905.00639, 2019.

\bibitem{Marcel}
S.~Marcel, M.~S. Nixon, and S.~Z. Li, \emph{Handbook of Biometric
  Anti-Spoofing: Trusted Biometrics Under Spoofing Attacks}.\hskip 1em plus
  0.5em minus 0.4em\relax Springer Publishing Company, Incorporated, 2014.

\bibitem{forbes}
\BIBentryALTinterwordspacing
{Davey Winder}. (2019) {Hackers Claim ‘Any’ Smartphone Fingerprint Lock Can
  Be Broken In 20 Minutes}. [Online]. Available:
  \url{{https://www.forbes.com/sites/daveywinder/2019/11/02/smartphone-security-alert-as-hackers-claim-any-fingerprint-lock-broken-in-20-minutes/\#35a3540d6853}}
\BIBentrySTDinterwordspacing

\bibitem{cc}
\BIBentryALTinterwordspacing
{heise online}. (2013) {iPhone 5s Touch ID hack in detail}. [Online].
  Available:
  \url{https://www.heise.de/multimediadatei/iPhone-5s-Touch-ID-hack-in-detail-1965628.html}
\BIBentrySTDinterwordspacing

\bibitem{8585605}
I.~{Goicoechea-Telleria}, A.~{Garcia-Peral}, A.~{Husseis}, and
  R.~{Sanchez-Reillo}, ``Presentation attack detection evaluation on mobile
  devices: Simplest approach for capturing and lifting a latent fingerprint,''
  in \emph{2018 International Carnahan Conference on Security Technology
  (ICCST)}, Oct 2018, pp. 1--5.

\bibitem{livdet2013}
L.~{Ghiani}, D.~{Yambay}, V.~{Mura}, S.~{Tocco}, G.~L. {Marcialis}, F.~{Roli},
  and S.~{Schuckcrs}, ``Livdet 2013 fingerprint liveness detection competition
  2013,'' in \emph{2013 International Conference on Biometrics (ICB)}, June
  2013, pp. 1--6.

\bibitem{8608133}
E.~{Marasco}, S.~{Cando}, L.~{Tang}, L.~{Ghiani}, and G.~L. {Marcialis}, ``A
  look at non-cooperative presentation attacks in fingerprint systems,'' in
  \emph{2018 Eighth International Conference on Image Processing Theory, Tools
  and Applications (IPTA)}, Nov 2018, pp. 1--6.

\bibitem{Chingovska2019}
I.~Chingovska, A.~Mohammadi, A.~Anjos, and S.~Marcel, \emph{Evaluation
  Methodologies for Biometric Presentation Attack Detection}.\hskip 1em plus
  0.5em minus 0.4em\relax Cham: Springer International Publishing, 2019, pp.
  457--480.

\bibitem{matsu}
T.~Matsumoto, H.~Matsumoto, K.~Yamada, and S.~Hoshino, ``{Impact of artificial
  "gummy" fingers on fingerprint systems},'' in \emph{Optical Security and
  Counterfeit Deterrence Techniques IV}, R.~L. van Renesse, Ed., vol. 4677,
  International Society for Optics and Photonics.\hskip 1em plus 0.5em minus
  0.4em\relax SPIE, 2002, pp. 275 -- 289.

\bibitem{Cao2016HackingMP}
K.~Cao and A.~K. Jain, ``Hacking mobile phones using 2 d printed
  fingerprints,'' 2016.

\bibitem{GHIANI2017110}
L.~Ghiani, D.~A. Yambay, V.~Mura, G.~L. Marcialis, F.~Roli, and S.~A.
  Schuckers, ``Review of the fingerprint liveness detection (livdet)
  competition series: 2009 to 2015,'' \emph{Image and Vision Computing},
  vol.~58, pp. 110 -- 128, 2017.

\bibitem{latentjain}
A.~Jain and J.~Feng, ``Latent fingerprint matching,'' \emph{IEEE transactions
  on pattern analysis and machine intelligence}, vol.~33, pp. 88--100, 01 2011.

\bibitem{gonzlezsoler2019fingerprint}
L.~J. González-Soler, M.~Gomez-Barrero, L.~Chang, A.~Pérez-Suárez, and
  C.~Busch, ``Fingerprint presentation attack detection based on local features
  encoding for unknown attacks,'' 2019.

\bibitem{8756285}
Y.~{Zhang}, D.~{Shi}, X.~{Zhan}, D.~{Cao}, K.~{Zhu}, and Z.~{Li},
  ``Slim-rescnn: A deep residual convolutional neural network for fingerprint
  liveness detection,'' \emph{IEEE Access}, vol.~7, pp. 91\,476--91\,487, 2019.

\end{thebibliography}

\end{document}